# BPMN4CAI: A BPMN Extension for Modeling Dynamic Conversational AI

**Research Paper**

Björn-Lennart Eger[1], Daniel Rose[1], and Barbara Dinter[1]

[1] Chemnitz University of Technology, Chair of Business Information Systems - Business Process and Information Management, Chemnitz, Germany
{bjoern-lennart.eger, daniel.rose, barbara.dinter}@wirtschaft.tu-chemnitz.de

**Abstract.** Conversational AI systems, such as chatbots and virtual assistants, are becoming increasingly important to digital business processes. However, the established Business Process Model and Notation (BPMN) standard faces challenges when representing dynamic, context-sensitive interactions. This paper addresses this methodological and practical research gap by developing a standard-compliant BPMN extension (BPMN4CAI). Using Design Science Research methodology, this paper develops an approach that systematically extends existing BPMN elements and incorporates specialized components. The applicability and relevance of the BPMN4CAI framework are demonstrated and evaluated through a case study. The results show that the BPMN4CAI extension facilitates adaptive decision-making processes, robust context management, and transparent interactions for Conversational AI within business processes.

## 1 Introduction

Chatbots and virtual assistants are established in numerous business areas, performing tasks that range from customer communication to automating internal processes. Originally designed as rule-based systems, these technologies have evolved into context-aware agents thanks to significant advancements in the field of Artificial Intelligence (AI), which are capable of conducting human-like dialogues and responding flexibly to conversational situations. Particularly, Conversational AI, which relies on natural language processing and adaptive learning, unlocks new potential for enhancing efficiency and customer focus (Zillmann et al. 2024; Buxmann et al. 2024; Car et al. 2020).

In practice, Conversational AI is increasingly used to automate dialogue-based processes, such as customer service, appointment scheduling, or information provision (Bitkom e. V. 2020). However, this poses new challenges for the systematic integration into existing process landscapes. Business Process Model and Notation (BPMN), the established modeling language in business process management, provides a standardized notation for modeling processes (Drescher 2017; Dumas et al. 2021), but has been primarily designed for deterministic processes. In contrast, modern Conversational AI introduces characteristics like non-deterministic behavior, context-dependent decision logic, and situational escalations, which are not adequately captured by traditional BPMN methods (Braun et al. 2014). For instance, dynamic dialogue paths, flexible reactions to unclear inputs, or adaptive information queries are difficult to represent in the notation.

Although initial research already explores approaches to integrating Conversational AI into BPMN (López et al. 2019; Lins, Melo et al. 2021), there is a lack of a formally grounded and standard-compliant modeling framework that meets the specific requirements of dialogue-oriented AI systems. Existing approaches often remain technology-centered or address isolated use cases without advancing the underlying process notation. This leads to two main research gaps: (1) methodologically, there is an absence of standard-compliant, theoretically founded concepts for depicting context-sensitive, dialogue-oriented systems in BPMN, and (2) practically, the integration of hybrid actors, such as chatbots, into existing process notations has been inadequately addressed. These gaps are derived in Section 2.3 and motivate the conceptual development of a standard-compliant extension. In this context, this paper aims to develop a methodologically sound extension of BPMN (BPMN4CAI) that enables the adequate integration of Conversational AI as a hybrid process actor. This leads to the following central research question:

*How can Conversational AI be integrated into the BPMN notation?*

In the following, Section 2 will first outline the theoretical foundations of Conversational AI and BPMN. Subsequently, Section 3 describes the methodological approach along the design-science-research logic. In Section 4, the conceptual development of the BPMN4CAI extension is carried out, whose applicability and benefits are demonstrated in a proof of concept in Section 5 and evaluated in Section 6. Section 7 summarizes key insights and outlines perspectives for further research.

# 2 Fundamentals and State of Research

## 2.1 Conversational AI: Fundamentals and Potentials

Conversational AI encompasses advanced chat and voice-based systems that enable human-like interactions using techniques from Machine Learning and Natural Language Processing (NLP) (Mariani et al. 2023; Khatri et al. 2018). Unlike simple, rule-based chatbots such as ELIZA (Weizenbaum 1966), advanced systems learn from large datasets and dynamically adapt to varying conversational contexts. Developments like GPT-4 (Buxmann et al. 2024) or BERT (Devlin et al. 2019) capture contextual relationships over longer dialogues and produce solution-oriented responses in near real-time.

The deployment of such systems has demonstrated effectiveness across diverse business processes, such as automating orders or in customer service (Rizk et al. 2020;

Wecke 2024). Interest in dialogue-oriented automation is also growing in areas like healthcare or administration (Milne-Ives et al. 2020; Hafner et al. 2024). However, technical challenges, such as unpredictable outcomes, and ethical challenges, like bias in training data, persist (Kieslinger et al. 2024; Seufert et al. 2023). The systematic integration into formal process models could increase transparency and controllability. Considering these potentials and challenges, the role of Conversational AI in the context of business processes will be discussed in more detail below.

### 2.2 Conversational AI as a Hybrid Process Actor

Examining the fundamental properties of Conversational AI prompts the question of how these systems function in real business processes. Traditionally, business processes distinguish between human actors with high flexibility and technical systems with high efficiency (Hilmer 2016; Dumas et al. 2021). Conversational AI merges the adaptability of human decision-making with the operational efficiency of technical systems. Through adaptive, probabilistic learning mechanisms, these systems can respond flexibly to linguistic inputs, allowing them to exhibit hybrid, human-like behavior (Preuss et al. 2023; Niederer et al. 2022). Conversational AI presents an attractive option for communication-intensive processes due to its ability to reduce manual effort and ensure constant availability. Existing modeling approaches, like BPMN, are predominantly tailored for deterministic processes and thus fall short in capturing the dynamic, dialogue-oriented essence of Conversational AI (Dumas et al. 2021; Drescher 2017). This discrepancy highlights the need for specialized modeling concepts that enable the integration of dialogue-oriented, hybrid actors into formal process models (Seufert et al. 2023; Barton et al. 2022).

### 2.3 Current State of Research

While the fundamental properties of Conversational AI and its hybrid role as a process actor have already been outlined, literature analysis shows that existing studies predominantly focus on converting existing BPMN process models into chatbot-based dialogue systems or on supporting users in process execution through Conversational Agents (López et al. 2019; Lins, Melo et al. 2021). For example, Lins, Melo et al. (2021) present an approach where process-supporting Conversational Agents (Process-Aware Conversational Agent (PACA)) guide users through predefined process paths via voice-based interactions. Similarly, López et al. (2019) introduce a method for automatically generating interactive chatbots from formal BPMN models that provide users with flexible assistance in process execution. These studies currently serve as pivotal references in this domain.

Additional studies explore specific facets of AI components within business processes, such as the orchestration and control of AI-based services (Wolters et al. 2020), proactive process monitoring (prescriptive monitoring) using AI (Zeltyn et al. 2022), or the use of generative AI models for automated process modeling from unstructured text data (Vidgof et al. 2023). Complementary practical case studies, including the application of generative models for process automation, highlight important technical potentials and challenges (Lins, Nascimento et al. 2023).

Overall, it is evident that research on integrating Conversational AI into BPMN remains highly fragmented and predominantly addresses specific technological issues. In particular, standardized methodological concepts are lacking, which would facilitate the systematic and consistent formal integration of hybrid actors, such as Conversational Agents, into existing process models. These identified research gaps constitute the foundational basis for the current investigation. Two central research gaps thus emerge:

1. **Methodologically**: There is a lack of model-theoretically grounded concepts for integrating context-sensitive, conversational AI functions into BPMN, which consider both formal requirements and the interactivity of language-based systems.
2. **Practically**: The role of hybrid actors like chatbots is not yet systematically represented in established process notations.

These research gaps motivate the development of BPMN4CAI, a standard-compliant extension, which enables the adequate integration of dialogue-oriented AI actors.

## 3 Methodological Approach

This paper adopts an artifact-centric approach using the Design Science Research (DSR) methodology as described by Peffers et al. (2007). The aim is to develop an extension of BPMN that enables the integration of Conversational AI as a hybrid process actor—an approach designed to meet both theoretical and practical requirements. The procedure explicitly adheres to the six steps described by Peffers et al. (ibid.): problem identification, objective definition, artifact development, demonstration, evaluation, and communication.

The methodological approach comprises three consecutive steps:

- **Requirements analysis and artifact development**: Based on a systematic literature review, key requirements for integrating dynamic, context-sensitive AI functions into existing BPMN models were identified (see Steps 1–3 according to Peffers et al. (ibid.)). The research was conducted in the databases Scopus, IEEE Xplore, and SpringerLink using topic-specific search terms (e.g., “Conversational AI,” “BPMN,” “process modeling”). Contributions with a methodological reference to the process modeling of dialogue-oriented AI systems were included; purely technical studies lacking a notation reference were excluded. The resulting requirements served as the foundation for the development of the specific extension BPMN4CAI, which enables the modeling of dialogue-based processes and automated tasks.
- **Design and prototypical implementation**: The model was designed based on the derived requirements and practically implemented in a proof of concept framework (see Step 4). Using a realistic scenario, it was demonstrated how the conceptual requirements can be realized in a concrete model extension. The focus was on assessing the feasibility, comprehensibility, and semantic expressiveness of the approach, rather than on empirical generalizability.

- **Demonstration and conceptual evaluation**: The extension's suitability was assessed through a specific use case from insurance consulting (see Steps 5–6). This conceptual evaluation served to illustrate how effectively key requirements can be modeled as well as to identify the benefits and limitations of the approach. The transferability to other application contexts will be explored in future research.

This methodological approach enables the development of a prototype-tested model that serves as a solid foundation for further research in the field of AI-supported business processes. Through the iterative development and reflection process, the aim is to increase both replicability and connectivity for research and practice.

# 4 Concept of the BPMN Extension

## 4.1 Objective Definition and Conceptual Foundations

The aim of this paper is to develop an artifact-centric approach that enables the integration of Conversational AI into BPMN. At the core of this approach is the creation of a hybrid process actor that supports both automated, data-driven decisions and dialogue-based, natural language interactions.

**Functional Categories of Conversational AI** To systematically capture the requirements for integrating Conversational AI into business processes, the following functional categories are derived:

- **User Interaction**: Conversational AI processes natural language requests and delivers relevant information in real-time (Buxmann et al. 2024; Rizk et al. 2020; Milne-Ives et al. 2020).
- **Context Preservation**: By continuously capturing and managing the dialogue context, AI enables personalized and consistent interactions (Florindi et al. 2024; Moiseeva et al. 2020; Niederer et al. 2022).
- **Automated Decisions**: AI makes decisions based on data analyses and can dynamically map alternative process paths (Bolliger et al. 2024; Luo et al. 2022; Buxmann et al. 2024).
- **Data Query and Management**: For informed decisions, AI accesses external data sources, whose dynamic retrieval and management are integral parts of modern business processes (Dean et al. 2023; Saha et al. 2024; Hafner et al. 2024).
- **Process Execution, Error Handling, and Escalation**: Besides actively performing tasks, such as initiating workflows, AI must be able to detect errors and, if necessary, escalate complex requests to human handlers to ensure transparency and regulatory compliance (Koohborfardhaghighi et al. 2023; Parker et al. 2024; Barton et al. 2022; European Parliament & European Council 2016; European Parliament & European Council 2024).

**Derivation and Deduction of Modeling Requirements** The previously presented functional categories form the basis for deriving specific modeling requirements (as detailed in Table 1):

1. **Representation of natural language interactions**: It must be possible to adequately represent both the dynamic dialogue context and variable user inputs (cf. REQ B.1).
2. **Mapping context-sensitive, dynamic decisions**: AI-based decision-making should control alternative, flexible process paths in real-time (cf. REQ C.1 and REQ C.2).
3. **Dynamic retrieval of external data**: It must be possible to retrieve external data sources in a context-sensitive manner and integrate them into the process (cf. REQ D.1).
4. **Transparent escalation mechanisms**: In the case of inappropriate AI decisions, a clear handover to human handlers must occur, including the transfer of relevant context information (cf. REQ E.1 to REQ E.3).

A summary of these requirements can be found in Table 1.

To thoroughly derive the need for extension, an equivalence analysis was conducted, systematically analyzing whether the derived requirements can be adequately modeled using the BPMN standard. The analysis shows that the BPMN standard does not sufficiently cover the modeling of ongoing dialogues, dynamic decision processes, context-sensitive data integration, and transparent escalations. For example, Service Tasks and User Tasks only capture static interactions and do not offer mechanisms for representing dialogical, context-aware processes (cf. REQ B.1, REQ B.2). The decision logic of standard gateways is also bound to fixed rules and does not allow dynamic path control based on AI results (REQ C.1, REQ C.2). Similarly, with data integration: Data Objects secure the predefined data flow but do not model adaptive logic for context-driven retrieval of external information, unlike dynamic, context-sensitive data streams (REQ D.1, REQ D.2). Finally, the standard lacks the means to specifically hand over escalations with complex context to human handlers (REQ E.1–E.3). The equivalence analysis thus confirms the necessity of a targeted extension of the BPMN standard to adequately address the special requirements of Conversational AI in business processes.

**Table 1.** Summary of Requirements for BPMN Integration of Conversational AI

| REQ | Core Requirement |
|---|---|
| **A) Modeling Efficiency and Transparency** | |
| REQ A.1 | Ensure compatibility with the existing BPMN standard. |
| REQ A.2 | Ensure transparent and easily understandable modeling. |
| REQ A.3 | Ensure uniform representation and terminology according to BPMN standards. |
| REQ A.4 | Provide technically precise foundations for implementation. |
| **B) Process Integration of Conversational AI** | |
| REQ B.1 | Representation of natural language interactions in the process. |
| REQ B.2 | Continuous capture of the dialogue context over multiple steps. |
| REQ B.3 | Mapping context-related AI interactions in the specific process context. |
| REQ B.4 | Ensure capture and processing of speech inputs and outputs. |
| **C) Decision-Making and Flexibility** | |
| REQ C.1 | Map dynamic decisions based on AI results. |
| REQ C.2 | Flexible alternative process paths with varying AI results. |
| **D) Data Integration and Context Management** | |
| REQ D.1 | Enable dynamic retrieval and integration of external data sources. |
| REQ D.2 | Ensure management of context and process data over multiple steps. |
| REQ D.3 | Represent access to documents and databases to support AI applications. |
| **E) Traceability and Escalation** | |
| REQ E.1 | Transparent representation of AI decisions in the process. |
| REQ E.2 | Enable escalation to human handlers for complex requests. |
| REQ E.3 | Map clear transfer of context and process data during escalations. |

### 4.2 Conception and Development of the BPMN4CAI Extension

The equivalence analysis presented in Section 4.1 shows that the BPMN standard reaches its limits in the areas of dynamic, context-sensitive interactions and decision-making. To close these gaps, an extension approach was developed that does not necessitate the creation of entirely new elements but focuses on the targeted enhancement of existing BPMN components. Classic elements—such as *Service Tasks, User Tasks, Events, and Gateways*—are enhanced with specific attributes and supplemented with specialized derivatives as needed.

The extension is achieved by enriching standard BPMN components with attributes such as *naturalLanguageProcessing* (activation of natural language processing), *contextManagement* (management of ongoing dialogue context), as well as *system_message_prefix* and instructions (transmission of control information). This approach ensures compatibility with the existing standard (cf. REQ A.1–A.4) and allows for the strategic reuse of established modeling elements. The technical extension is carried out in compliance with the extension mechanism described in the BPMN 2.0 specification via <*extensionElements*> (cf. OMG 2011, p. 44). Consequently, the new elements can be integrated into existing BPMN-compliant tools and processed by them while maintaining compliance with the standard.

Based on this foundation, specialized elements were derived to address the extended requirements of Conversational AI:

- **Conversational Task**: Extension of the Service Task to map interactive, natural language dialogues. In addition to naturalLanguageProcessing and contextManagement, parameters including inputType (text or speech input), model (used AI model), and temperature (response creativity) are used to configure the interaction.
- **Function Calling Task**: Supports the dynamic invocation of external interfaces. Attributes such as apiEndpoint, requestPayload, and responseType enable precise modeling of data access and systemic actions.
- **Data Retrieval Task**: Serves the context-sensitive retrieval of external data. dataSource, queryParameters, and expectedResponseType specify source, query, and format.
- **Information Management Task**: Manages the storage and updating of context data. storeData, dataFormat, and updateContext ensure consistent data across process steps.
- **Human Escalation Event**: Allows forwarding to human actors. escalationReason, assignedTo, and contextData facilitate comprehensive information transfer.
- **AI Decision Gateway**: Facilitates dynamic decision-making based on attributes such as decisionLogic, threshold, and decisionCriteria.

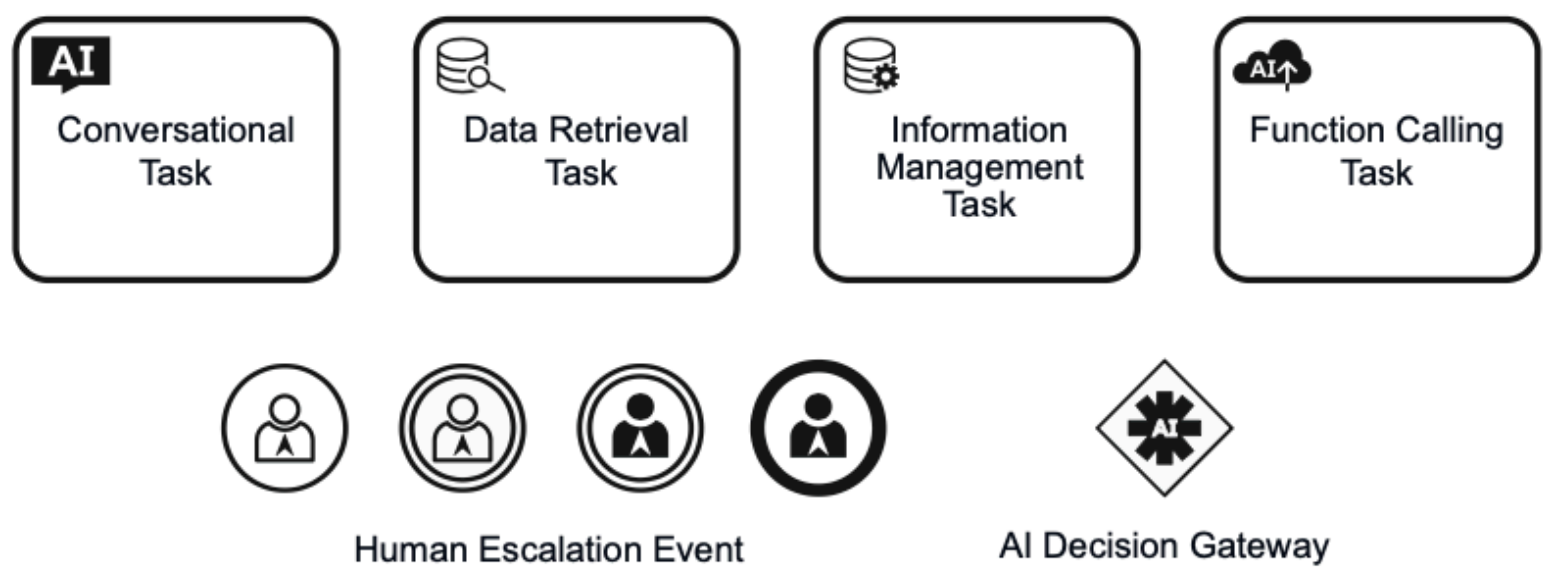

**Figure 1.** Graphical Representation of the Elements of the BPMN4CAI Adaptation (self-created representation; icons: self-created)

The conception of the BPMN extension combines the modification of existing elements with the introduction of BPMN4CAI-specific components. The approach ensures compatibility with the BPMN standard (REQ A.1–A.4), enables the mapping of context-sensitive interactions (REQ B.1–B.4), supports dynamic decisions (REQ C.1, C.2), allows external data access as well as robust context management (REQ D.1–D.3), and ensures structured escalation (REQ E.1–E.3).

With the help of the BPMN extension mechanism, the BPMN4CAI elements can be seamlessly integrated into existing modeling tools without changing the basic structure of BPMN. The extension follows a BPMN+X approach, where the modeled framework additionally exists as an XML instance of an extended XML Schema Definition (XSD) schema[1]. This enables technical validation and potential further processing of the models in BPMN-compatible environments (OMG 2011).

```
1  <task id="Activity_0dpu93s" name="Present recommended products">
2    <extensionElements>
3      <ext:ConversationalTask
4        naturalLanguageProcessing="true"
5        contextManagement="true"
6        model="gpt-3.5-turbo"
7        temperature="0.7"
8        instructions="Present suitable product recommendations to the customer based
             on their specifications. [...]"
9        chat_history_limit="20" />
10   </extensionElements>
11 </task>
```

**Listing 1.1.** XML Representation of the Conversational Task for Product Presentation

Embedding is achieved through the <*extensionElements*> structure of the BPMN specification. Listing 1.1 demonstrates this with an example of a Conversational Task.

The use of the <*extensionElements*> mechanism corresponds to the method provided in the BPMN specification for extending existing elements (ibid.) and thus ensures the standard-compliant integration of additional information for AI-based process execution.

The XML representation serves as a machine-readable basis for the subsequent implementation of Conversational Agents and illustrates the integration of the extended modeling into existing tools and workflows.

In the following sections, the approach will be practically demonstrated and evaluated to illustrate the technical implementability as well as the modeling and practical benefits of the BPMN4CAI extension.

# 5 Demonstration of the BPMN Extension

After conceptualizing and describing the BPMN4CAI extension, we now proceed to the practical demonstration of its application. The aim is to demonstrate how the developed extension elements incorporate dynamic, context-sensitive AI interactions into business processes. As an example, a consultation process in the insurance industry is modeled, where a Conversational AI agent engages in a dialogue with a customer, requests information, makes data-driven decisions, and escalates to human caseworkers when necessary.

[1] https://gitlab.hrz.tu-chemnitz.de/-/snippets/245

In the depicted process (Figure 2), the use of a Conversational AI agent is demonstrated, managing customer inquiries and providing tailored insurance recommendations. The modeled process includes the following steps:

1. **Initiation of dialogue by the customer**: The interaction begins with a natural language request processed by the Conversational Task. This element replaces the standard User Task as it manages not only language input and output but also the ongoing dialogue context (contextManagement). This explicitly highlights in the model that it is a dialogue-based, context-aware interaction—a differentiation that would only be implicit or possible through external documentation with standard BPMN.
2. **Retrieval of context-sensitive data**: Relevant information, such as customer data or product recommendations, is retrieved from external sources through the Data Retrieval Task. Unlike generic Service Tasks, this element allows for precise specification of the data source (dataSource) and query parameters (queryParameters), which is crucial for both readability and implementation.
3. **Dynamic decision-making on the recommendation path**: The AI Decision Gateway determines the further course—either direct recommendation or escalation. Unlike classic gateways, this decision is not based on fixed rules but on data-driven evaluation logics, controllable via attributes like decisionCriteria and threshold.
4. **Escalation to human advisors**: If the agent cannot make a well-founded recommendation, a handover is executed through the Human Escalation Event. This includes not only the escalation logic but also structured context data (contextData), ensuring clear traceability and adherence to regulatory requirements, such as those from the GDPR or the AI Act.

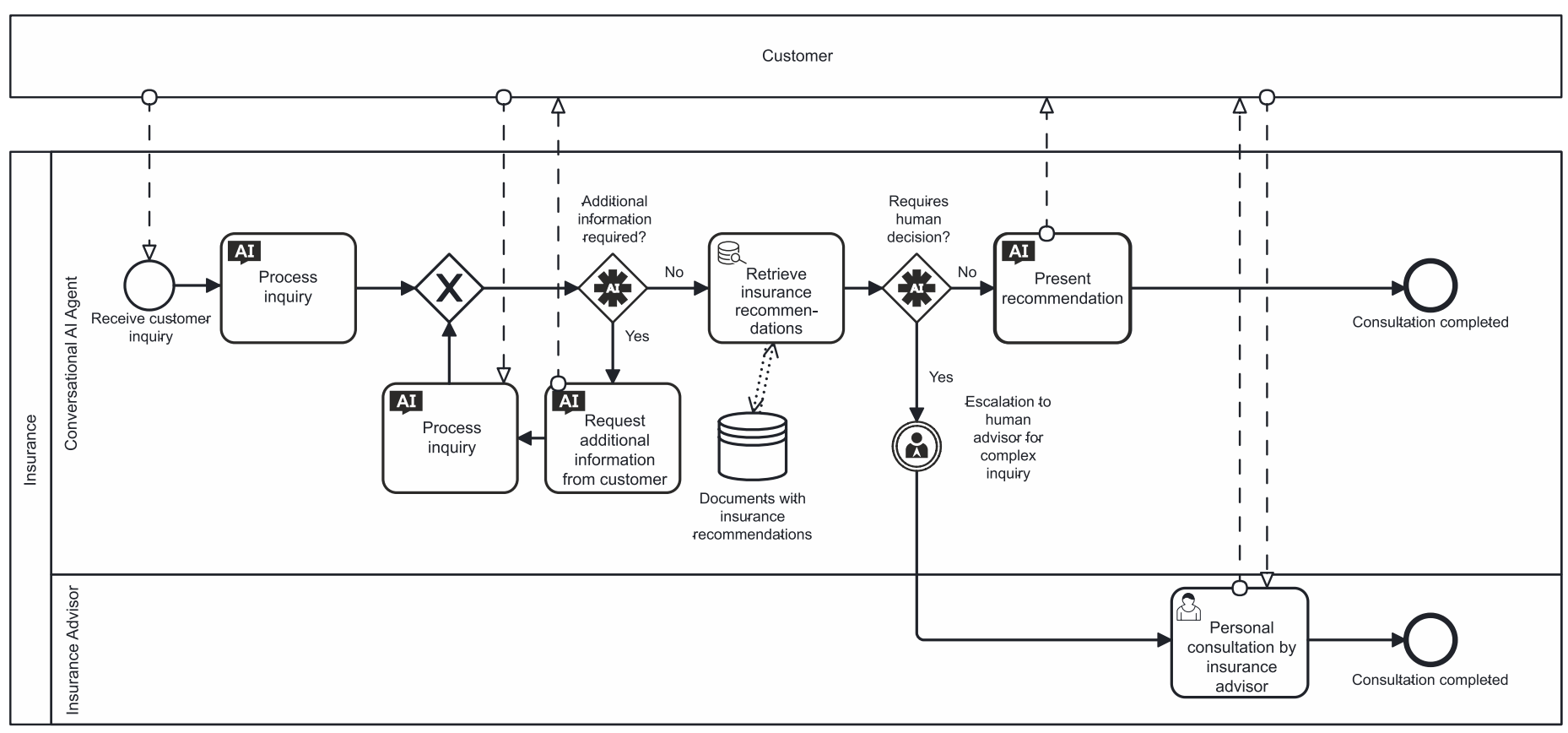

**Figure 2.** BPMN4CAI Model: Consultation on Recommended Insurances

The modeling demonstrates that existing BPMN elements such as User Task or Service Task are inadequate for depicting dialogue-oriented, context-sensitive AI interactions. On the one hand, there is a lack of distinctive visualization to differentiate dialogical agent processes from standard, rule-based tasks. On the other hand, key technical configuration aspects—such as language model control or context data management—cannot be directly integrated into the models. BPMN4CAI bridges this gap through a combination of visually distinguishable elements and structured attributes, which improve both model understanding and form a semantic bridge to implementation.

This use case should be interpreted as a proof of concept that exemplarily demonstrates how the developed BPMN4CAI elements can be used in a realistic business process, rather than as a generalizable case study. The goal is to show how effectively the extension can model fundamental requirements, its semantic expressiveness, and technical connectivity. A systematic evaluation of the transferability to other application areas is the subject of future work.

## 6 Evaluation and Discussion

The BPMN4CAI extension was validated through a proof of concept evaluating the integration of dialogue-based, context-sensitive interactions in a prototypical advisory process in the insurance industry. The graphical modeling and technical transformation into a valid BPMN-XML schema demonstrate both the applicability and implementability of the new elements in common BPMN-compatible tools.

The evaluation shows that the developed elements address key requirements for the modeling of Conversational AI: The *Conversational Task* allows for the explicit representation of language-based interactions; *Function Calling Task* and *Data Retrieval Task* conceptually separate external function calls and data access; the *AI Decision Gateway* facilitates dynamic decision-making based on AI-generated outcomes, and the *Human Escalation Event* facilitates the handover to human actors with context transfer.

At the same time, the evaluation also revealed limitations: The modeling of alternative paths in dynamic, non-deterministic processes remains constrained. While simple decision logics can be well represented, multi-layered, adaptively changed process paths quickly reach the expressive limits of the BPMN syntax. Adequate representation of such processes would necessitate additional mechanisms, such as rule sets or domain-specific extensions.

The explicit representation of context transfer during escalations also remains a challenge: Although data transfers can be modeled using BPMN data objects, there is a lack of standard-compliant and intuitive visualization, especially to create understandable handover processes for less technically versed modelers. Additionally, some of the new task types exhibit functional overlap, which may unnecessarily complicate modeling practice.

These limitations indicate that although the BPMN4CAI extension offers essential functionalities, it does not yet comprehensively address all facets of complex, AI-sup-ported interactions. Future research should explore approaches for more intuitive modeling of non-deterministic decision-making, clearer differentiation of task types, and improved visualization of escalation logic, alongside systematic domain verification.

## 7 Conclusion and Outlook

This paper aims to integrate Conversational AI as a hybrid process actor into the BPMN notation. For this purpose, the BPMN4CAI extension was developed, enabling the dynamic and context-sensitive representation of natural language interactions, flexible, non-deterministic decision processes, and the retrieval of external data sources. The practical application, demonstrated through an example in the advisory process of an insurance consultation, confirms that the developed extension elements—such as the Conversational Task, the Function Calling Task, the Data Retrieval Task, the Human Escalation Event, and the AI Decision Gateway—provide significant advantages over traditional BPMN models.

The evaluation further demonstrates that the BPMN standard reaches its limits in areas such as the representation of dynamic decisions and the management of continuous dialogue contexts. Through systematic equivalence testing, it was methodically confirmed that targeted extensions are necessary to meet the requirements of modern AI-supported business processes. The integration of new elements was validated initially through a specific use case. However, further investigations are needed regarding the transferability to other industries and more complex scenarios.

Reflection on the technical implementation is also necessary. Although the implementation of the BPMN4CAI extension—such as the transformation into an XML schema—was demonstrated as a proof of concept, future work should evaluate more closely the interoperability of the extended elements in a broader application framework.

Furthermore, initial explorations in the field of Agentic AI, where AI systems proactively trigger independent actions, offer promising perspectives. These synergies between Conversational AI and agent-based AI open up additional possibilities for the optimization and automation of business processes, which should serve as a starting point for further research.

Overall, this approach seeks to contribute to the integration of Conversational AI into standardized business process models. The BPMN4CAI extension offers a robust foundation for comprehensively modeling dynamic and context-sensitive interactions—although further research is needed regarding the transferability to different industries and the complete technical integration. In the future, these aspects should be explored in depth, while also further investigating interfaces to process mining, multi-agent systems, and legal and ethical issues.

## References

Barton, M.-C., & Pöppelbuß, J. (2022). Prinzipien für die ethische Nutzung künstlicher Intelligenz. *HMD Praxis der Wirtschaftsinformatik*, *59*(2).

Bitkom e. V. (2020). *Jedes vierte Unternehmen in Deutschland nutzt Chatbots*. Retrieved from https://www.bitkom.org/Presse/Presseinformation/Jedes-vierte-Unternehmen-in-Deutschland-nutzt-Chatbots (visited on 03/15/2025)

Bolliger, D., & Lienhard, S. (2024). Dialoge, die verkaufen: Mit Conversational Automation zur Kundenbindung. In N. Hafner & S. Hundertmark (Eds.), *Kundendialog-Management. Wertstiftende Kundendialoge in Zeiten der digitalen Automation*. Springer Gabler.

Braun, R., & Esswein, W. (2014). Extending BPMN for modeling resource aspects in the domain of machine tools. *WIT Transactions on Engineering Sciences*, *87*.

Buxmann, P., Glauben, A., & Hendriks, P. (2024). Die Nutzung von ChatGPT in Unternehmen: Ein Fallbeispiel zur Neugestaltung von Serviceprozessen. *HMD Praxis der Wirtschaftsinformatik*, *61*(2).

Car, L. T., Dhinagaran, D. A., Kyaw, B. M., Kowatsch, T., Joty, S., Theng, Y. L., & Atun, R. (2020). Conversational agents in health care: Scoping review and conceptual analysis. *Journal of Medical Internet Research*, *22*(8).

Dean, M., Burke, L., Parkinson, B., Leavy, S., O'Sullivan, D., & Greene, D. (2023). Chatpapers: An AI chatbot for interacting with academic research. In *2023 31st Irish Conference on Artificial Intelligence and Cognitive Science (AICS)*.

Devlin, J., Chang, M.-W., Lee, K., & Toutanova, K. (2019). BERT: Pre-training of deep bidirectional transformers for language understanding. In *Proceedings of the 2019 Conference of the North American Chapter of the Association for Computational Linguistics: Human Language Technologies, Volume 1 (Long and Short Papers)*.

Drescher, A. (2017). *Modellierung und Analyse von Geschäftsprozessen. Grundlagen und Übungsaufgaben mit Lösungen*. De Gruyter Oldenbourg.

Dumas, M., La Rosa, M., Mendling, J., & Reijers, H. A. (2021). Einführung in das Geschäftsprozessmanagement. In M. Dumas, M. La Rosa, J. Mendling, & H. A. Reijers (Eds.), *Grundlagen des Geschäftsprozessmanagements*. Springer Berlin Heidelberg.

European Parliament & European Council. (2016). *Regulation (EU) 2016/679 of the European Parliament and of the Council of 27 April 2016 on the protection of natural persons with regard to the processing of personal data and on the free movement of such data, and repealing Directive 95/46/EC (General Data Protection Regulation)*.

European Parliament & European Council. (2024). *Regulation (EU) 2024/1689 of the European Parliament and of the Council of 13 June 2024 laying down harmonised rules on artificial intelligence and amending Regulations (EC) No 300/2008, (EU) No 167/2013, (EU) No 168/2013, (EU) 2018/858, (EU) 2018/1139 and (EU) 2019/2144 and Directives 2014/90/EU, (EU) 2016/797 and (EU) 2020/1828 (Artificial Intelligence Act)*.

Florindi, F., Fedele, P., & Dimitri, G. M. (2024). A novel solution for the development of a sentimental analysis chatbot integrating ChatGPT. *Personal and Ubiquitous Computing*.

Hafner, N., & Hundertmark, S. (Eds.). (2024). *Kundendialog-Management. Wertstiftende Kundendialoge in Zeiten der digitalen Automation*. Springer Gabler.

Hilmer, C. (2016). Prozessmanagement und indirekte Bereiche. In C. Hilmer (Ed.), *Prozessmanagement in indirekten Bereichen. Empirische Untersuchung und Handlungsempfehlungen*. Springer Fachmedien Wiesbaden.

Khatri, C., Hedayatnia, B., Venkatesh, A., Nunn, J., Pan, Y., Liu, Q., Song, H., Gottardi, A., Kwatra, S., Pancholi, S., Cheng, M., Huang, Q., Misra, A., Lamba, A., Einolghozati, A., Shi,

S., Beirami, A., Eghbali, A., Hancock, S., ... Hakkani-Tur, D. (2018). Alexa Prize — State of the art in conversational AI. *AI Magazine*, *39*(3).

Kieslinger, K., & Nierobisch, K. (2024). Wenn der Chatbot weiß, wo es lang geht – Ethische Fragen und mögliche Kriterien zum Einsatz von KI-gestützten Beratungsettings. *Zeitschrift für Weiterbildungsforschung*, *47*(1).

Koohborfardhaghighi, S., De Geyter, G., & Kaliner, E. (2023). Unlocking the power of LLM-based question answering systems: Enhancing reasoning, insight, and automation with knowledge graphs. In *International Conference on Intelligent Systems Design and Applications*. Springer.

Lins, L. F., Melo, G., Montoya, E., Ceravolo, P., Damiani, E., & Tavares, G. M. (2021). PACAs: Process-aware conversational agents. In *International Conference on Business Process Management*. Springer.

Lins, L. F., Nascimento, N., Pessoa, M., Barbosa, E., Lima, L., Oliveira, K., Perkusich, A., & Tavares, G. M. (2023). Comparing generative chatbots based on process requirements: A case study. In *2023 IEEE International Conference on Big Data (BigData)*. IEEE.

López, A., Mateo, F., Sánchez-Ferreres, J., Carmona, J., & Padró, L. (2019). From process models to chatbots. In *Advanced Information Systems Engineering: 31st International Conference, CAiSE 2019, Rome, Italy, June 3–7, 2019, Proceedings 31*. Springer.

Luo, B., Lau, R. Y., Li, C., & Si, Y. W. (2022). A critical review of state-of-the-art chatbot designs and applications. *Wiley Interdisciplinary Reviews: Data Mining and Knowledge Discovery*, *12*(1).

Mariani, M. M., Hashemi, N., & Wirtz, J. (2023). Artificial intelligence empowered conversational agents: A systematic literature review and research agenda. *Journal of Business Research*, *161*.

Milne-Ives, M., Lam, C., De Cock, C., Van Velthoven, M. H., & Meinert, E. (2020). The effectiveness of artificial intelligence conversational agents in health care: Systematic review. *Journal of Medical Internet Research*, *22*(10).

Moiseeva, A., Santhanam, S., Ackermann, C., Weiss, R., Scheurenbrand, J., Stier, C., Runkler, T., & Tresp, V. (2020). *Multipurpose intelligent process automation via conversational assistant*. Retrieved from http://arxiv.org/pdf/2001.02284.

Niederer, T., Schloss, D., & Christensen, N. (2022). Designing context-aware chatbots for product configuration. In *International Workshop on Chatbot Research and Design*. Springer.

Object Management Group. (2011). *Business Process Model and Notation (BPMN), Version 2.0*. Retrieved from https://www.omg.org/spec/BPMN/2.0/PDF.

Parker, M. J., Chao, J., Fontana, F. A., Garcia-Lopez, E., Lavoie, A., Lee, L., McBride, C., & Perera, T. (2024). A large language model approach to educational survey feedback analysis. *International Journal of Artificial Intelligence in Education*.

Peffers, K., Tuunanen, T., Rothenberger, M. A., & Chatterjee, S. (2007). A design science research methodology for information systems research. *Journal of Management Information Systems*, *24*(3).

Preuss, P., Horstmann, M., & Kaper, N. (2023). Kombination von Chatbots und Robot Process Automation – Disruption in Kundenservice-Centern? *Banking & Innovation 2022/2023*.

Rizk, Y., Awad, M., & Turcotte, E. W. (2020). *A unified conversational assistant framework for business process automation*. Retrieved from http://arxiv.org/pdf/2001.03543.

Saha, B., & Saha, U. (2024). Enhancing international graduate student experience through AI-driven support systems: A LLM and RAG-based approach. In *2024 International Conference on Data Science and Its Applications (ICoDSA)*. IEEE.

Seufert, S., & Meier, C. (2023). Hybride Intelligenz: Zusammenarbeit mit KI-Assistenzsystemen in wissensintensiven Bereichen. *HMD Praxis der Wirtschaftsinformatik*, *60*(6).

Vidgof, M., Bachhofner, S., & Mendling, J. (2023). Large language models for business process management: Opportunities and challenges. In *International Conference on Business Process Management*. Springer.

Wecke, B. (2024). *Wachstum durch den Einsatz von Generativer KI*. Springer Fachmedien Wiesbaden.

Weizenbaum, J. (1966). ELIZA — a computer program for the study of natural language communication between man and machine. *Communications of the ACM*, *9*(1).

Wolters, B., & Köhne, F. (2020). *bpmn.ai: Process patterns to orchestrate your AI services in business processes*. Camunda. Retrieved from https://camunda.com/blog/2020/11/bpmn-ai-process-patterns-to-orchestrate-your-ai-services-in-business-processes/#group2 (visited on 09/15/2024)

Zeltyn, S., Nelken, R., Pelecanos, J., Ouaknine, J., Cuenca, E., & Huang, M. (2022). *Prescriptive process monitoring in intelligent process automation with chatbot orchestration*. Retrieved from http://arxiv.org/pdf/2212.06564.

Zillmann, M., & Hahn, G. (2024). *Generative AI – Von der Innovation bis zur Marktreife. Wo stehen Unternehmen im deutschsprachigen Raum bei der Nutzung von Generative AI?* Lünendonk & Hossenfelder GmbH. Retrieved from https://www.protiviti.com/sites/default/files/2024-09/luenendonk_generative-ai_studie_protiviti-de.pdf (visited on 09/15/2024)